# Crash Severity Prediction Using Deep Learning Approaches: A Hybrid CNN-RNN Framework

**Sahar Koohfar**
Department of Civil Engineering
University of Texas at San Antonio, San Antonio, Texas, United State of America
Email: Sahar.koohfar@my.utsa.edu

**Amit Kumar**
**Department io Civil Engineering**
University of Texas at San Antonio, San Antonio, Texas, United State of America
Email: Amit.Kumar@utsa.edu

**ABSTRACT**

Accurate and timely prediction of crash severity is crucial in mitigating the severe consequences of traffic accidents. Accurate and timely prediction of crash severity is crucial in mitigating the severe consequences of traffic accidents. In order to provide appropriate levels of medical assistance and transportation services, an intelligent transportation system relies on effective prediction methods. Deep learning models have gained popularity in this domain due to their capability to capture non-linear relationships among variables. In this research, we have implemented a hybrid CNN-RNN deep learning model for crash severity prediction and compared its performance against widely used statistical and machine learning models such as logistic regression, naïve bayes classifier, K-Nearest Neighbors (KNN), decision tree, and individual deep learning models: RNN and CNN. This study employs a methodology that considers the interconnected relationships between various features of traffic accidents. The study was conducted using a dataset of 15,870 accident records gathered over a period of seven years between 2015 and 2021 on Virginia highway I-64. The findings demonstrate that the proposed CNN-RNN hybrid model has outperformed all benchmark models in terms of predicting crash severity. This result illustrates the effectiveness of the hybrid model as it combines the advantages of both RNN and CNN models in order to achieve greater accuracy in the prediction process.

**Keywords:** Crash Severity, RNN, CNN, Deep Learning, Machine Learning.

## INTRODUCTION

Traffic infrastructures serve as vital components of people's daily routines worldwide. Maintaining road safety is essential for the overall welfare of society, encompassing social and economic aspects. However, the occurrence of fatalities and injuries resulting from motor vehicle accidents is a significant global public health concern. In the United States, vehicle crashes have resulted in substantial economic and societal losses. The reports a concerning statistic, projecting that approximately 42,915 individuals lost their lives in motor vehicle traffic crashes in 2021. This represents a significant 10.5% increase compared to the 38,824 fatalities recorded in 2020 [1]. The prevention of accidents and the reduction of severe accidents are pivotal objectives within the transportation profession.

In order to improve traffic safety, it is crucial to gain a comprehensive understanding of the factors that contribute to the severity of injury. Having this understanding allows policymakers to predict and anticipate future crashes, as well as their severity, considering certain circumstances. To identify the factors that contribute to crash severity, it is essential to use accurate prediction models. Consequently, numerous studies have been conducted to examine the contributing factors associated with these accidents. Researchers have employed various types of models to better understand the underlying causes and risk factors. The objective of these models is to identify the key variables and relationships that affect the occurrence and severity of crashes. Models for predicting crash severity can generally be divided into two categories: statistical and machine learning models.

The statistical model establishes a mathematical relationship between the explanatory variables and crashes severity based on uncertainty distribution assumptions and hypothesis tests. Such models can help isolate the effects of explanatory variables on crash severity. Several statistical learning models have been used to analyze crash injury severity, including probit, logit, ordered probit, and mixed probit.

In their research, Chen et al. employed probit models to examine the factors influencing injury severity in truck-involved crashes [2]. Similarly, Abdel-Aty conducted a study focusing on driver injury severity using multinomial logit (MNL) and ordered probit models [3].Moreover, Kim et al. employed mixed logit models to identify the factors that contribute to motor vehicle collisions involving pedestrians [4]. In another study, Fan et al. conducted a comparative analysis between the MNL model and the ordered logit (OL) model to assess their performance in identifying and examining the influence of contributing variables on crash severity levels at highway-rail crossings [5].

Discrete choice models, such as the multinomial logit and ordered logit models, are commonly used in crash severity analysis. However, these models have certain assumptions and pre-defined relationships between the dependent and explanatory variables, which can restrict their application when these assumptions are violated. While these models are simple to implement and require less training, they may struggle to capture the nonlinearity of crash severity and the complex relationships among contributing factors.

As a result of these limitations, a variety of machine learning methods have been developed for modeling crash severity. These advanced techniques offer the potential to effectively capture the nonlinear and hidden relationships between factors that influence crash severity outcomes. Contrary to traditional statistical models, these approaches do not rely on strict assumptions and are capable of learning complex patterns directly from the data. As a result, they have the advantage of being flexible and can be adapted to different data distributions and variables interactions without requiring any substantive assumptions to be made in the past.

Li et al. applied support vector machine (SVM) model for crash severity prediction and compared its performance with OP models. The results indicated that SVM outperformed the OP model in terms of accuracy when analyzing crash injury severity [6]. AlMamlook et al. compared the performance of different machine learning algorithms for predicting traffic accident severity and found that the Random Forest algorithm was the most accurate, with an accuracy of 75.5%. The other algorithms tested, including Logistic Regression, Naïve Bayes, and AdaBoost, had accuracies ranging from 73.1% to 74.5% [7].

in a study conducted by Beshah et al., researchers applied three data mining techniques, namely decision tree, Naïve Bayes, and K-nearest neighbors (KNN), to establish connections between recorded road characteristics and accident intensity in Ethiopia [8]. Based on the findings of this study, a set of rules was developed in order to improve road safety in Ethiopia. A study conducted by Nandurge et al. used K-means clustering algorithm and association rule mining as data mining techniques to discover the factors associated with traffic accidents[9]. Another study utilized the K-means clustering algorithm to identify the most frequently occurring accident-prone areas and the key factors associated with these accidents[10].

The decision tree is another machine learning method that has been used in the forecasting of crash severity. In a recent study, Kong et al. investigated and compared the performance of decision trees, artificial neural networks (ANNs), and random forests in analyzing crash severity using data from the Traffic Accident Analysis System (TAAS) [11]. Similarly, Alkheder et al compared the probit algorithm with ANNs combined with k-means clustering for predicting traffic crashes. In their study, they found that ANNs were more accurate at predicting crash severity than probit algorithms. A neural network method has been shown to be superior to other methods for predicting crash severity, making it a popular method among crash severity prediction researchers [12].

Deep learning–based techniques have been applied to a wide range of applications in the transportation industry, including traffic flow analysis and accident hotspot prediction. Koohfar and Dietz employed multi-scale temporal transformers to capture recurring patterns and similarities in traffic flow across different temporal contexts (e.g., hours, days, and weeks). Their study demonstrated that incorporating these dynamic temporal similarities enhanced the transformer's forecasting accuracy by up to 50% [13, 14]. In their other work, Koohfar and Dietz enhanced the forecasting capability of foundation models by removing smoothly correlated noise commonly present in time-series data such as traffic flow. This approach improved traffic flow prediction accuracy by approximately 10%. Their effort highlights the potential of deep learning models to address complex, transportation-related problems [15]. Deep learning-based techniques have also been applied to a variety of applications within the transportation industry, including traffic flow analysis and accident hotspot prediction [16, 17]. These techniques have demonstrated promising results in improving the accuracy and efficiency of transportation-related prediction models. Sameen et al. used the Recurrent Neural Network (RNN) model to predict injury severity of traffic accidents in Malaysia. By using a grid search method, the researchers were able to determine the optimal network architecture for suboptimal network parameters.

The RNN model outperformed the Multilayer Perceptron (MLP) and Bayesian Logistic Regression (BLR) models with a best validation accuracy of 71.77% [17]. Zheng et al. examined the effectiveness of convolutional neural networks (CNNs) for predicting road traffic crashes (RTCs). As part of the study, the authors compared the performance of the proposed CNN model with nine commonly used statistical and machine learning models, including KNN, decision trees, support vector machines, and neural networks (NN). According to the comparison results, the CNN model performed better than the other models in predicting RTC severity, with a F1score of 84% [18]. Although many machine learning techniques have been employed for predicting crash severity, few studies have attempted to compare the performance of recurrent neural networks and convolutional neural networks with each other. Moreover, To the best of the authors' knowledge, no prior research has been conducted on the prediction of crash severity using a hybrid RNN-CNN model.

The purpose of this study is to address the limitations of the existing literature by comparing commonly used machine learning and deep learning models for crash severity prediction and developing a novel hybrid RNN-CNN model.

Consequently, this study seeks to provide insights into the effectiveness of different machine learning models for predicting crash severity and contribute to the development of models that are more accurate and efficient. CNN models perform well at identifying spatial features in images, such as road and weather conditions, while RNN models capture temporal dependencies in sequential data, such as crash events. The combination of a RNN and a CNN can provide a more comprehensive and accurate understanding of crash severity than either model alone. Comparing RNN and CNN performance can also aid in identifying the most effective ML models for predicting crash severity by providing insight into their strengths and weaknesses. This work could contribute to the development of more accurate and efficient methods for predicting crash severity, ultimately resulting in improved road safety.

## METHODS

For this paper, the crashes data was obtained from the Virginia Department of Transportation and Development (VOTD). The dataset comprises15,840 crashes that took place on Virginia Highway I-64 between 2015 and 2021. With a length of 299 miles (481 km), Interstate 64 (I-64) traverses the state of Virginia in an east-west direction, cutting through the center of the state from West Virginia all the way to the Hampton Roads region. Figure 1 illustrates the precise route of the I-64 highway.

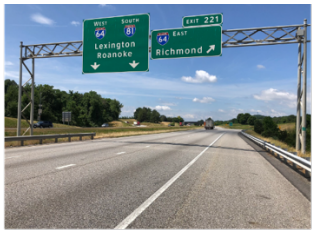

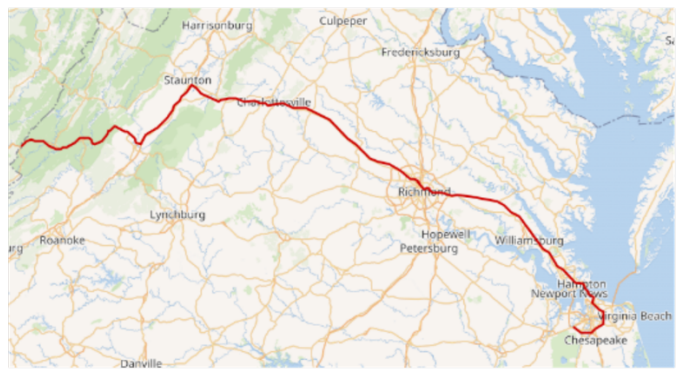

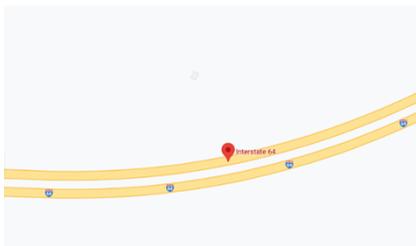

**Figure 1. I-64 highway. Source:(googlemap.com)**

Traffic congestion poses a significant challenge in modern times and has contributed to an increase in road accidents. This paper addresses the problem of crash severity prediction by utilizing various statistical and machine learning techniques. Specifically, decision tree (DT), logistic regression (LR), naïve bayes classifier, k-nearest neighbors (KNN), as well as deep learning models including recurrent neural networks (RNN), convolutional neural networks (CNN), and the hybrid CNN-RNN are employed.

**Data**

When examining crash severity, the dependent variable consists of discrete categories representing different levels of severity. Previous studies have commonly classified crash severity into three levels: property damage only (PDO), injury, and fatality [19, 20]. These severity levels can be treated as either nominal or ordinal variables, allowing for flexibility in the selection of modeling techniques.

In the current study, crash severity was classified into four distinct levels: fatal (K), non-fatal (O), suspected serious injury (A), and suspected minor injury and possible injury (B, C). Figure 1 provides an overview of the severity levels of crash-related injuries. Within the dataset analyzed, there were 81 fatal crashes (K), 755 crashes with suspected serious injury (A), 3307 crashes with minor injury and possible injury (B, C), and 11697 crashes with no apparent injury (0). The distribution of crash severity levels within the dataset is visually represented in figure 2.

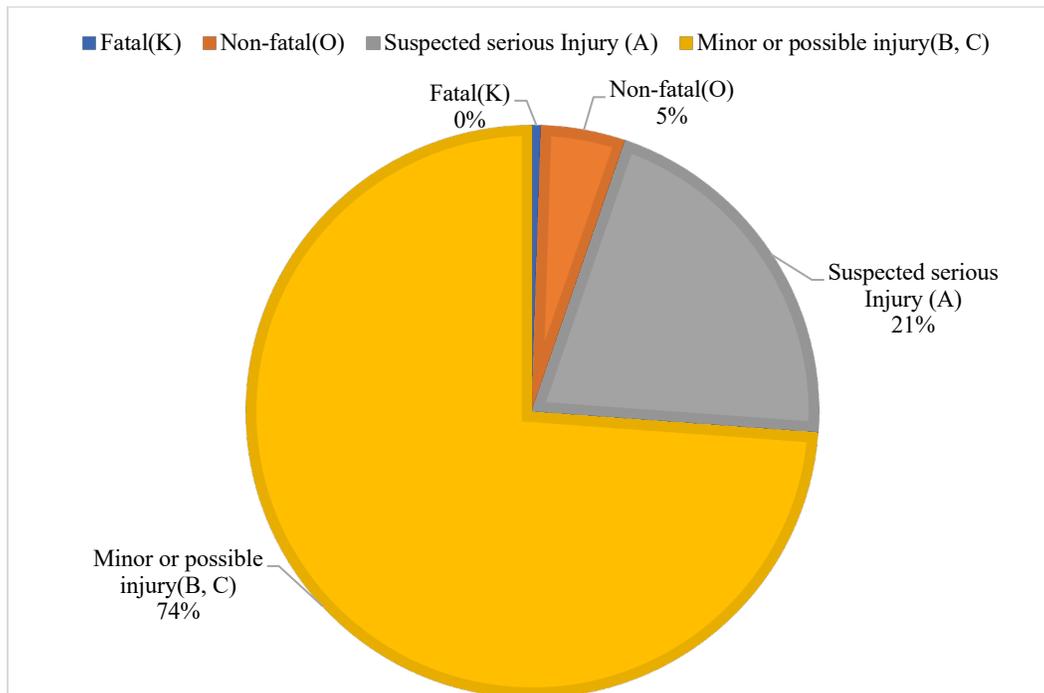

**Figure 2. Crash severity level distribution**

The dataset used in this study consisted of 45 variables, which included various details related to crashes such as a unique crash ID, year, date, time of the crash, and other relevant information about crash-related factors, environmental conditions, roadway characteristics, and vehicle features. During the initial data analysis, irrelevant variables and those with missing values were filtered out. In this study, specific features were chosen as potential factors that might affect the severity of the crashes. These features included lighting conditions, belt condition, drug condition, location of the first harmful indecent, traffic control device, road type, and whether the crash occurred during nighttime. Additionally, the number of vehicles involved in the crash, pedestrian action at time of the accident, and the type of area where the accident took place were also considered as independent variables in the analysis.

Table 1 presents a statistical summary of the variables that were included and examined in this study, providing an overview of their characteristics and distribution.

**Table 1. Statistical summary of variables**

| Variable | code | Variable Sub-Classification | Max | Min | Mean | STD |
|---|---|---|---|---|---|---|
| Crash Severity | 0 | Property Damage Only (No Apparent Injury) | 3 | 0 | 1.0321 | 0.3047 |
| | 1 | Suspected Minor Injury, Possible Injury | | | | |
| | 2 | Suspected Serious Injury | | | | |
| | 3 | Fatality | | | | |
| Weather Condition | 1 | No Adverse Condition (Clear/Cloudy) | 11.00 | 1.00 | 1.8794 | 1.7028 |
| | 3 | Fog | | | | |
| | 4 | Mist | | | | |
| | 5 | Rain | | | | |
| | 6 | Snow | | | | |
| | 7 | Sleet/Hail | | | | |
| | 8 | Smoke/Dust | | | | |
| | 9 | Other | | | | |
| | 10 | Blowing Sand, Soil, Dirt, or Snow | | | | |
| | 11 | Severe Crosswinds | | | | |
| Light Condition | 1 | Dawn | 7.00 | 1.00 | 2.7113 | 1.1883 |
| | 2 | Daylight | | | | |
| | 3 | Dusk | | | | |
| | 4 | Darkness - Road Lighted | | | | |
| | 5 | Darkness - Road Not Lighted | | | | |
| | 6 | Darkness - Unknown Road   Lighting | | | | |
| | 7 | Unknown | | | | |
| Road Type | 1 | Two-Way, Not Divided | 3.00 | 0.00 | 0.5433 | 0.9340 |
| | 2 | Two-Way, Divided, Unprotected | | | | |
| | 3 | Median | | | | |
| | 4 | Two-Way, Divided, Positive | | | | |
| | 5 | Median Barrier<br>One-Way, Not Divided<br>Unknown | | | | |
| Location of First harmful Event | 0 | Not Provided | 9.00 | 1.00 | 1.4270 | 1.0582 |
| | 1 | On Roadway | | | | |
| | 2 | Shoulder | | | | |
| | 3 | Median | | | | |
| | 4 | Roadside | | | | |
| | 5 | Gore | | | | |
| | 6 | Separator | | | | |
| | 7 | In Parking Lane or Zone | | | | |
| | 8 | Off Roadway, Location Unknown | | | | |
| | 9 | Outside Right-of-Way | | | | |
| Traffic Control Device | 0 | Not Provided | 6.00 | 1.00 | 1.0408 | 0.4400 |

| | 1 | Yes - Working | | | | |
|---|---|---|---|---|---|---|
| | 2 | Yes - Working and Obscured | | | | |
| | 3 | Yes - Not Working | | | | |
| | 4 | Yes - Not Working and Obscured | | | | |
| | 5 | Yes - Missing | | | | |
| | 6 | No Traffic Control Device Present | | | | |
| Traffic Control Type | 0 | Not Provided | 17.00 | 1.00 | 6.2880 | 1.7800 |
| | 1 | No Traffic Control | | | | |
| | 2 | Officer or Flagger | | | | |
| | 3 | Traffic Signal | | | | |
| | 4 | Stop Sign | | | | |
| | 5 | Slow or Warning Sign | | | | |
| | 6 | Traffic Lanes Marked | | | | |
| | 7 | No Passing Lines | | | | |
| | 8 | Yield Sign | | | | |
| | 9 | One Way Road or Street | | | | |
| | 10 | Railroad Crossing with Markings and Signs | | | | |
| | 11 | Railroad Crossing with Signals | | | | |
| | 12 | Railroad Crossing with Gate and Signals | | | | |
| | 13 | Other | | | | |
| | 14 | Pedestrian Crosswalk | | | | |
| | 15 | Reduced Speed - School Zone | | | | |
| | 16 | Reduced Speed - Work Zone | | | | |
| | 17 | Highway Safety Corridor | | | | |
| Pedestrian Action | 0 | Not Provided | 3.00 | 0.00 | 0.0015 | 0.0495 |

| | 1 | Crossing at intersection with Signal | | | | |
|---|---|---|---|---|---|---|
| | 2 | Crossing at intersection again Signal | | | | |
| | 3 | Crossing at intersection no Signal | | | | |
| Alcohol Condition | 0 | No | 1.00 | 0.00 | 0.0368 | 0.18862 |
| | 1 | Yes | | | | |
| Drug Condition | 0 | No | 1.00 | 0.00 | 0.0060 | 0.07761 |
| | 1 | Yes | | | | |
| Young Driver Condition | 0 | No | 1.00 | 0.00 | 0.1882 | 0.3909 |
| | 1 | Yes | | | | |
| Belt Condition | 0 | No | 1.00 | 0.00 | 0.02714 | 0.16251 |
| | 1 | Yes | | | | |
| Night Condition | 0 | No | 1.00 | 0.00 | 0.2827 | 0.4503 |
| | 1 | Yes | | | | |
| Area Type | 0 | Rural | 1.00 | 0.00 | 0.7610 | 0.4246 |
| | 1 | Urban | | | | |
| Vehicle Count | | - | 9.0 | 1.0 | 1.958 | 0.8987 |

*Data preprocessing*

This study has identified several features as potential determinants of crash severity in the preliminary data analysis. The categorical variables are transformed into binary dummy values as a preprocessing step in order to evaluate the impact of each individual factor on the severity of the crash. As a result of this flattening process, it is possible to assess more accurately the effect each variable has on crash severity.

The dataset was subsequently divided into two separate subsets: a test set and a training set. 25% of the data was allocated to the test set, and 75% to the training set. To facilitate the integration of the categorical variables into the deep learning models used in this study, the categorical variables were converted into time series. Deep learning models such as RNN and CNN are capable of processing and analyzing time series data more efficiently.

*Feature Selection*

In this study, an ensemble-based feature selection method called extra trees classifier was utilized to enhance the classification performance by reducing the data processing workload. The extra trees classifier is a variation of ensemble methods based on decision trees, which combines multiple decision trees to perform feature selection and make predictions. By employing this approach, the study aimed to identify the most influential features that have a substantial impact on predicting the target variable.

The selected features, which had higher importance scores, were then used as inputs in the machine learning models for crash severity prediction. These key features were included in these models, which enabled them to make more accurate predictions. Table 2 represents the top-ranking features based on their feature weight, providing insights into the variables that carry the most weight in the analysis. By examining the table, we have eliminated the pedestrian action, drug condition, and traffic control type variables, and opted to consider variables with importance scores greater than 0.025 for further analysis.

**Table 2.  Variable's importance score**

| Feature | Importance Score |
|---|---|
| First Harmful Event | 0.1675 |
| Vehicle Count | 0.1349 |
| Road Type | 0.1251 |
| Weather Condition | 0.1103 |
| Belt Condition | 0.1104 |
| Light Condition | 0.0864 |
| Alcohol Condition | 0.0652 |
| Area Type | 0.0486 |
| Traffic Control Device | 0.0415 |
| Young Driver Condition | 0.0355 |
| Night | 0.0261 |
| Traffic Control type | 0.0251 |
| Drug Condition | 0.0133 |
| Pedestrian Action | 0.0094 |

For better visualization, Figure 3 displays a bar plot representing the importance scores of each variable. In this plot, the eliminated variables are highlighted in red color, while the remaining variables are shown in green.

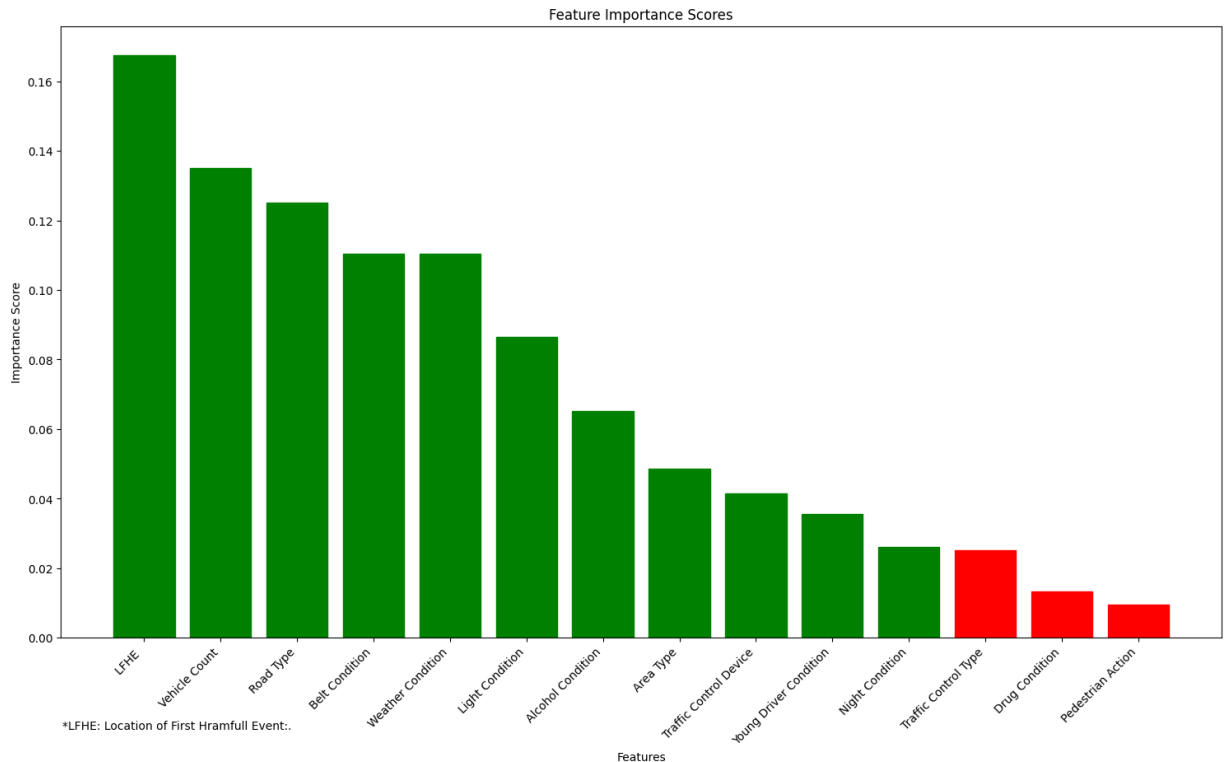

**Figure 3. Bar chart of importance score**

This paper utilizes statistical and deep learning models to predict crash severity levels. The models employed in this study encompass logistic regression, decision trees, K-Nearest Neighbor (KNN), naïve bayes, Recurrent Neural Networks (RNNs), Convolutional Neural Networks (CNNs), and the proposed Hybrid CNN-RNN model. The following sections present comprehensive explanations of each of these models.

*Logistic Regression*

Logistic regression is a statistical technique used to model the likelihood or probability of a categorical outcome based on one or more input variables. It is commonly used when the outcome of interest is binary, meaning it can take one of two values (e.g., true/false, yes/no). For scenarios with more than two possible outcomes, such as multiple categories or classes, multinomial logistic regression is employed. The logistic regression model calculates the probability of the outcome using a transformation called the logit function [21]. The formula for binary logistic regression can be expressed as:

$$\log\left(\frac{p}{(1-p)}\right) = \beta_0 + \beta_1 x_1 + \beta_2 x_2 + \cdots + \beta_r x_r \qquad (1)$$

Where $\log\left(\frac{p}{(1-p)}\right)$ represents the log-odds or logit transformation of the probability of the outcome, p is the probability of the outcome, and $\beta_0$, $\beta_1$, $\beta_2$, ..., $\beta_r$ are the coefficients associated with the input variables $x_1$, $x_2$, ..., $x_r$, respectively. The goal of logistic regression is to estimate the values of the coefficients ($\beta_0$, $\beta_1$, $\beta_2$, ..., $\beta_r$) that best fit the data and maximize the likelihood of observing the given outcomes based on the input variables. Using the logistic regression formula, the probability of individual "i" experiencing severity level "I" from the set of severity outcomes "J" can be expressed as follows [22]:

$$P_{iI} = \frac{e^{\hat{\beta} x_{iI}}}{\sum_{j=1}^{J} e^{\hat{\beta} x_{ij}}} \qquad (2)$$

Where x is the factor and $\hat{\beta}$ is the fixed coefficient for all individual.

Naïve Bayes *Classifier*

The naïve bayes classifier is a probabilistic classifier that utilizes the Bayesian theorem. It falls under the category of probability classifiers. It is referred to as " naïve" because it assumes a strong independence assumption among the input variables. Hence, it can also be called simple bayes or independence bayes. The primary goal of the naïve bayes classifier is to estimate the conditional probability of a specific class given the observed input variables. This is achieved by utilizing Bayes' theorem, which calculates the posterior probability by considering the prior probability and the likelihood [23]. By comparing the probabilities of different classes, the classifier can make predictions by selecting the class with the highest probability as the most likely outcome. The Bayesian theorem can be restated as follows:

$$P(C|X) = \frac{P(X|C) \times P(C)}{P(X)} \qquad (3)$$

where $P(C \mid X)$ represents the posterior probability of class $C$ given the input $X$. The term $P(X \mid C)$ denotes the likelihood of observing the input X given class $C$. $P(C)$ refers to the prior probability of class $C$, while $P(X)$ signifies the probability of observing the input $X$.

*Decision Tree*

The decision tree serves as a model that maps observations about an item to conclusions about its target value. The decision tree structure consists of root nodes, internal nodes, and leaf nodes. Each internal node represents a test condition on an attribute, while each branch represents the result of the test condition. On the other hand, each leaf node (or terminal node) is assigned a class label. The tree progressively splits the data based on these rules until reaching the leaf nodes, which represent the final classifications or outcomes. This approach offers several advantages, including its simplicity of implementation and its ability to provide intuitive explanations when compared to other classification algorithms [24].

*K- Nearest Neighbor (KNN)*

The K-nearest neighbors (KNN) algorithm was initially introduced by Cover and Hart [25]. It is a classification algorithm that aims to assign a class to an observation by considering the k closest observations in the feature space. The class that appears most frequently among the k nearest neighbors is assigned as the class of the new observation. When implementing KNN, two hyperparameters need to be specified: the number of k nearest neighbors and the choice of distance function used to measure the distance between observations in the feature space. The value of k determines the level of neighborhood considered when classifying an observation, while the distance function defines the metric used to calculate the distances between observations.

to calculate the distance between the data points and the existing clusters, and assign the data point to the closest cluster, Euclidean distance metric ($d$) is calculated as follow [26]:

$$d(X, \acute{X}) = \sqrt{(X_1 - \acute{X}_1)^2 + (X_1 - \acute{X}_1)^2 + \cdots + (X_n - \acute{X}_n)^2} \qquad (4)$$

Where $d(X, \acute{X})$ is the Euclidean distance between two data point $X$, $\acute{X}$.

*Recurrent Neural Network (RNN)*

A multitude of interconnected neurons comprise an artificial neural network (ANN), with numerous layers involved. The input layer takes in input values, the hidden layers modify those values, and the output layer generates the final output values. These layers are connected by weights. ANNs come in different classes and can be used for a wide range of purposes. A recurrent neural network (RNN) is a specific type of ANN that has recurrent connections [27]. This means that its current output is dependent on its previous output, as it stores and utilizes its previous output to compute its current output. Figure 3 depicts the structure of an RNN model.

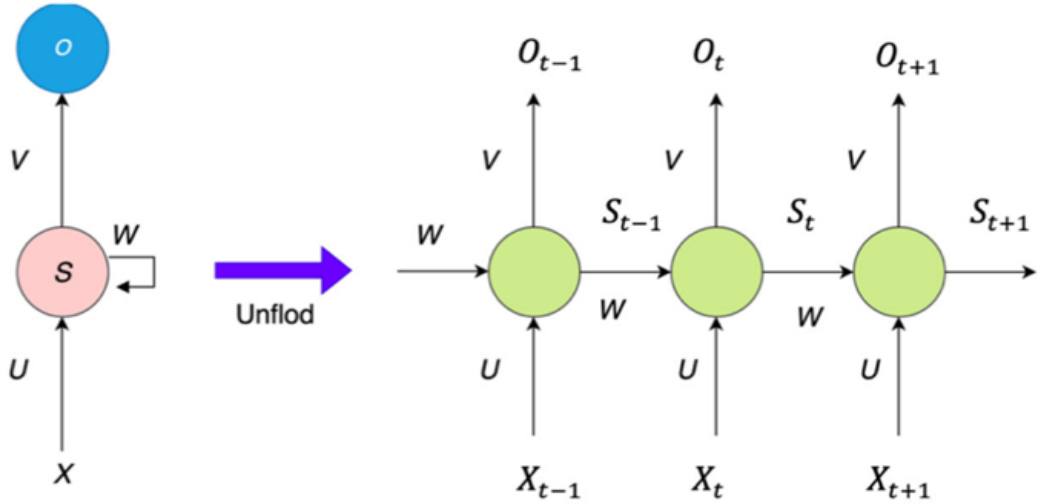

**Figure 1. RNN architecture, source:[28]**

Mathematically, for time t, the RNN formula is:

$$S_t = f\,(WS_{t-1} + Ux_t + b) \tag{5}$$

and

$$O_t = g\,(VS_t), \tag{6}$$

where $f$ and $g$ are the nonlinear activation functions (sigmoid functions), $S_t$ and $x_t$ represent the hidden layer and input layer, respectively, in the time step $t$, and $W$, $U$, and $V$ represent weight matrices. $O_t$ represents the output and b represents the bias.

### Convolutional Neural Network

A convolutional neural network (CNN) is an artificial neural network initially developed for processing image data and language processing applications [29]. Figure 4 shows a one-dimension CNN structure for a time-series forecasting model. It consists of an input layer, a convolutional layer, a pooling layer, a flattened layer, a fully connected layer, and an output layer. As can be seen in figure 4, either the convolution layer or the pooling layer is constructed of artificial neurons representing convolutional filters.

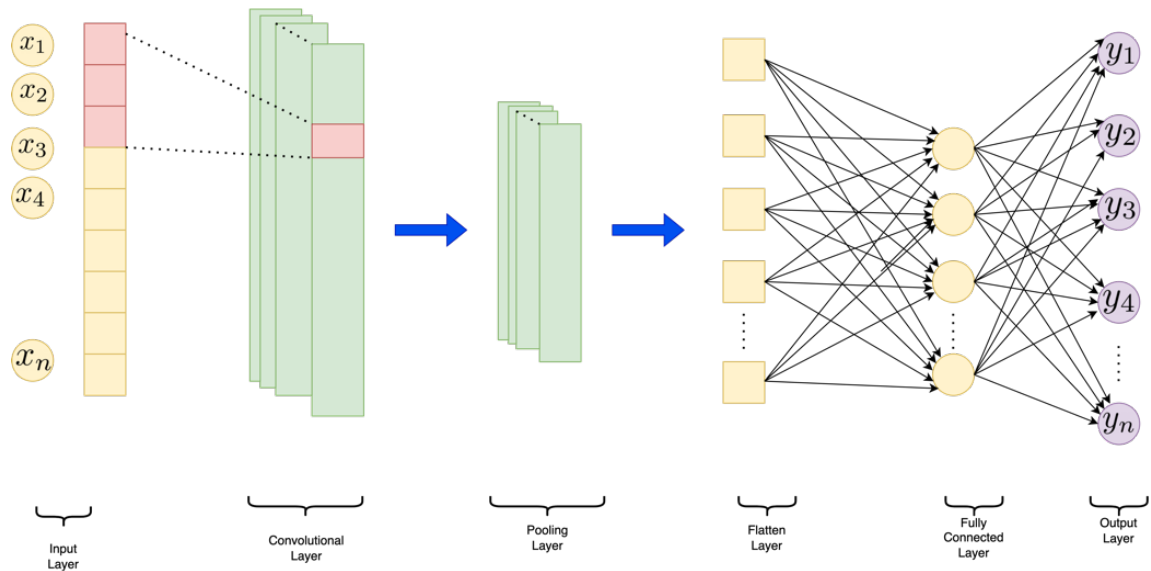

**Figure 2. CNN architecture. Source:[28]**

The convolutional layer reads the input and applies convolutional filtering to extract potential features. The pooling layer is used to reduce the input representation's spatial size and parameter count. A fully connected layer maps the features extracted by the network to classes or values.

*Hybrid CNN-RNN Model*

This section introduces a hybrid deep learning architecture designed for the detection and prediction of crash severity. The architecture consists of two primary modules: a convolutional neural used to extract features from input data, and a recurrent neural network used to predict the severity of the crash. The input data is first fed into the CNN module, where features are extracted. These features are then passed as input to the RNN module to predict the severity level of the crash. Figure 5 depicts the architecture of the proposed hybrid CNN-RNN model.

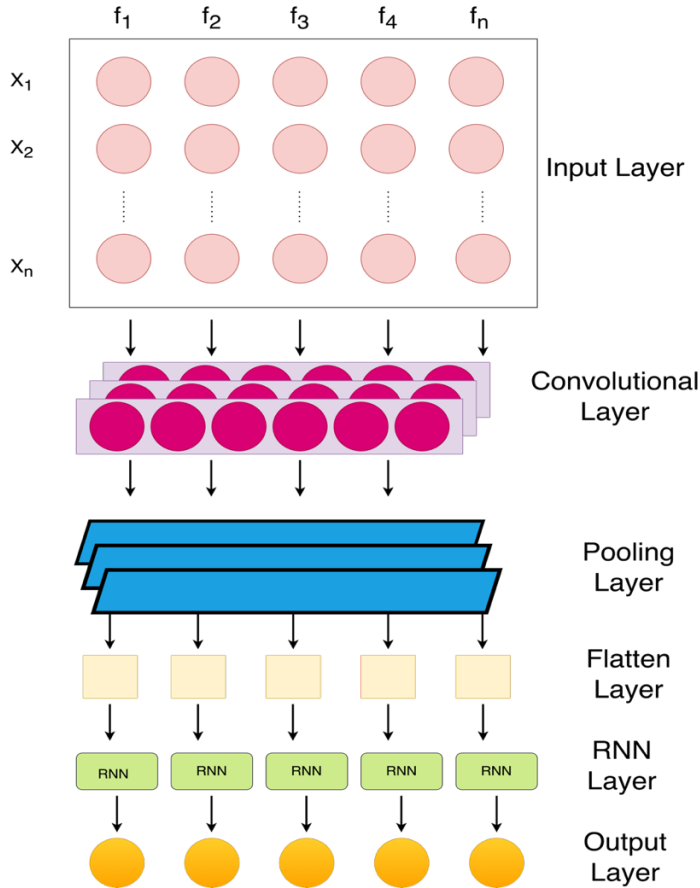

**Figure 3. Hybrid CNN-RNN architecture**

*Model Evaluation*

In this study, the effectiveness of different strategies was evaluated using widely accepted performance indicators, accuracy, precision, and recall. The formulas for selected evaluation indicators are presented below, where T.P. represents the true positive rate, T.N. is the true negative rate, F.P. denotes the false positive rate, and F.N. stands for the false negative rate.

$$\text{Accuracy} = \frac{\text{TP} + \text{TN}}{\text{TP} + \text{TN} + \text{FP} + \text{FN}} \qquad (7)$$

$$\text{Precision} = \frac{\text{TP}}{\text{TP} + \text{FP}} \qquad (8)$$

$$\text{Recall} = \frac{\text{TP}}{\text{TP} + \text{FN}} \qquad (9)$$

*Hyperparameter Optimization*

Hyperparameter optimization is a crucial step in machine learning model development. It involves tuning the parameters of the model to obtain the best possible performance. In our project, we have performed hyperparameter optimization by manually trying different combinations of hyperparameters and evaluating the performance of the model on a validation set. After several iterations, we have identified the best combination of hyperparameters that resulted in the highest accuracy. The details of the best hyperparameter combination and its corresponding performance metrics are presented in the table 3. This process of manual tuning allows us to find the optimal hyperparameters for our specific problem and dataset, resulting in improved model performance.

**Table 3. Hyperparameter values**

| Hyper Parameter | Best Value |
|---|---|
| Number of Layer | 6 |
| Model Dimension | 64 |
| Epoch | 50 |
| Activation Function | Relu, SoftMax |
| Drop Out | 0.002 |

*Addressing Data Imbalance*

The majority of the collected data consists of non-severe traffic crashes, due to the inherent characteristics of traffic accidents. Out of a total of 15,840 crash records gathered from 2015 to 2011, only 81 cases were classified as severe, accounting for a mere 1.00% of the dataset, as shown in figure 2. As a result, there would be a significant bias in favor of the non-severe class, which would be undesirable. This study utilized a generative model in order to achieve a balanced prediction of both severe and non-severe crashes. A generative model is developed by analyzing the existing data points and learning the characteristics and patterns of the minority group. After generating synthetic samples that closely represent the minority class, the classifier produces a balanced weight based on both classes' samples. The process continues until the number of samples in all classes is similar. Machine learning problems can be significantly improved by utilizing this method to predict rare events, such as severe crashes in this case. Figure 4 illustrates the resampling process employed to address the imbalanced class problem in crash severity data.

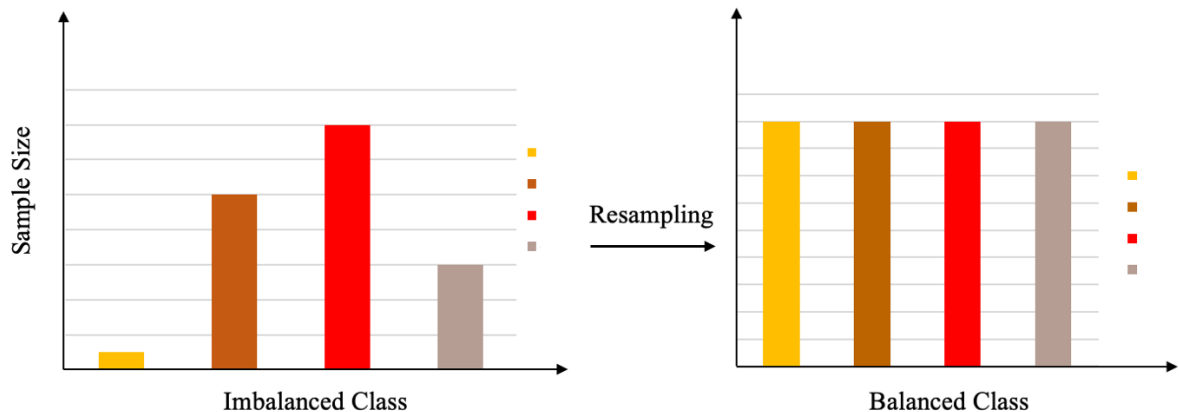

**Figure 4. Resampling of imbalanced class to balanced class.**

**RESULTS**

Table 4 presents the results obtained from an experiment conducted to predict crash severity using various statistical and machine learning models, including LR, naïve bayes, KNN, RNNs, CNNs, and RNN-CNNs. Multiple combinations of hyperparameters were evaluated and selected the best one for analysis. The findings clearly demonstrate the superiority of the deep learning models over statistical and machine learning models utilized in this study. Moreover, the hybrid RNN-CNN model outperformed the standalone RNN and CNN models across multiple evaluation metrics, such as test accuracy, precision, and recall.

Bu looking at the table 4, we can observe that the logistic regression model achieved an accuracy of 0.42, precision of 0.40, and recall of 0.42. The decision tree model performed slightly lower with an accuracy of 0.32, precision of 0.34, and recall of 0.32. KNN exhibited higher accuracy at 0.71, but its precision and recall values were relatively low at 0.27 and 0.25, respectively. The naïve bayes model achieved an accuracy of 0.69, precision of 0.40, and recall of 0.39. In general, it can be observed that the utilized machine learning and statistical models exhibited similar performance levels.

The RNN model achieved a test accuracy of 68%, whereas the CNN model achieved a slightly lower accuracy of 62%. However, the hybrid CNN-RNN model demonstrated superior performance by surpassing both models with an accuracy of 72%. In comparison to LR, DT, KNN, and naïve bayes, the proposed hybrid CNN-RNN model outperformed them with accuracy rates of 71%, 125%, 1%, and 4%, respectively. When considering precision and recall, the CNN-RNN model exhibited superiority with rates of 82%, 121%, 170%, 82%, 71%, 125%, 188%, and 84% compared to LR, DT, KNN, and the naïve bayes regression model, respectively.

CNN-RNN hybrid models outperformed individual RNN and CNN models in terms of precision. This indicates its improved ability to correctly identify relevant data and reduce false positives. Additionally, the hybrid model exhibited substantial improvements in recall, outperforming both the RNN and CNN models by 10% and 25% respectively. Furthermore, the RNN model also exhibited better recall performance compared to the CNN model.

Overall performance metrics were also noticeably improved with the hybrid model. It achieved a 4% increase in accuracy compared to the RNN model and an impressive 21% increase compared to the CNN model. The results of this study demonstrate the superiority of the CNN-RNN hybrid model in predicting crash severity levels and its potential for improving road safety.

**Table 4. Model's performance value**

| Model | Accuracy | Precision | Recall |
|---|---|---|---|
| **Logistic Regression** | 0.42 | 0.40 | 0.42 |
| **Decision Tree** | 0.32 | 0.34 | 0.32 |
| **KNN** | 0.71 | 0.27 | 0.25 |
| **Naïve Bayes** | 0.69 | 0.40 | 0.39 |
| **RNN** | 0.68 | 0.713 | 0.693 |
| **CNN** | 0.62 | 0.65 | 0.60 |
| **CNN-RNN** | 0.72 | 0.73 | 0.729 |

To provide a clearer understanding of the models' performance in terms of accuracy, recall, and precision, a bar chart figure is presented in Figure 7.

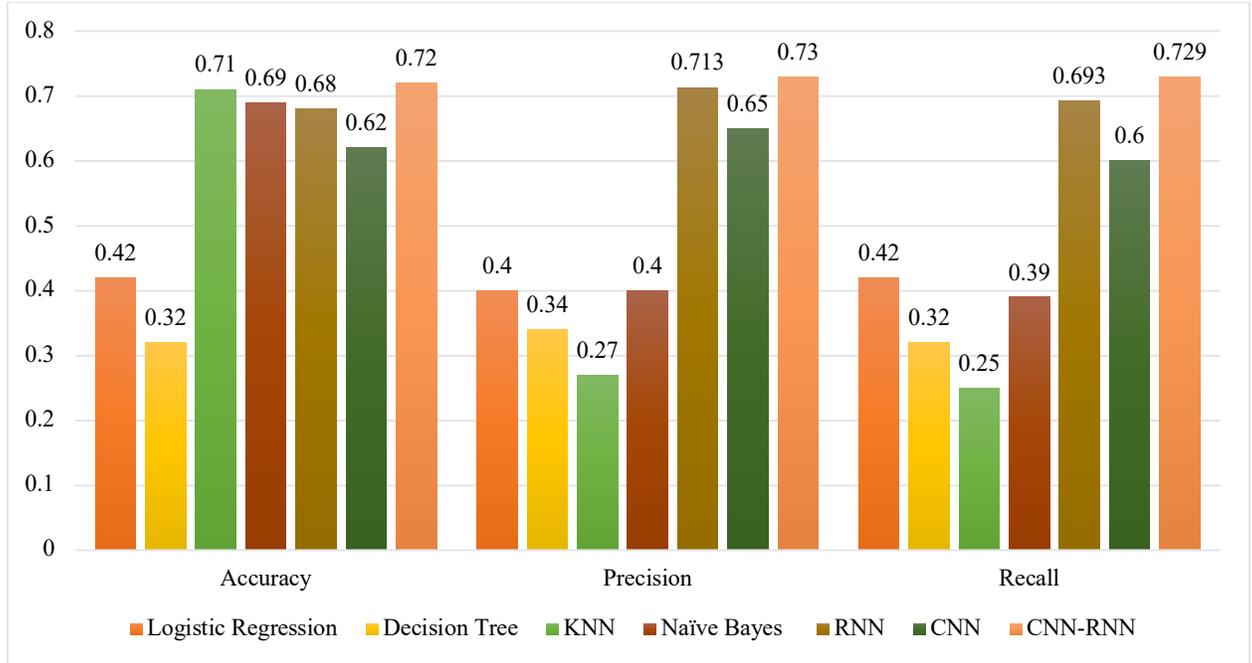

**Figure 7. Bar chart of model's performance**

**CONCLUSION**

Traffic collisions present a significant challenge for individuals who rely on road transportation for their daily activities. Various models, including statistical and machine learning approaches, have been utilized to forecast the severity of crashes and understand the factors influencing them. In this study, a hybrid deep learning model was proposed to address the problem of crash severity prediction. The performance of the proposed hybrid CNN-RNN model was compared to other commonly used models in this field, namely LR, DT, KNN, naïve bayes, RNN and CNN. The findings of this study indicate that the hybrid CNN-RNN model outperformed both models in terms of recall, accuracy, and precision.

The results of the study demonstrate that deep learning models exhibited superior performance compared to the statistical and machine learning models utilized in this research. This finding highlights the inherent strengths of deep learning models, particularly their capacity to effectively model complex and nonlinear data patterns. The ability of deep learning models to capture intricate relationships within the data contributes to their enhanced performance in this study, further underscoring their superiority over traditional machine learning and statistical models.

Furthermore, the performance comparison among the deep learning models indicates that the hybrid CNN-RNN model outperforms both the standalone RNN and CNN models. This improvement in accuracy can be attributed to the fact that the hybrid model combines the strengths of both RNN and CNN models. By combining both models, the hybrid model is able to effectively process both sequential and spatial data, resulting in a more accurate prediction. Additionally, the hyperparameter tuning process may have also contributed to the improved performance of the hybrid model. In conclusion, the CNN-RNN hybrid model appears to be promising for predicting crash severity. The hybrid approach demonstrates superior performance to individual RNN and CNN models by combining the strengths of RNNs and CNNs, and by meticulously tuning hyperparameters. For enhancing traffic safety, further research and validation are needed to confirm the generalizability and applicability of this hybrid model.

The study's findings have significant practical implications as they shed light on the comparative performance of machine learning and deep learning models in the context of road safety. In addition, the study proposes a hybrid deep learning model that can be applied effectively. In addition to assisting policy makers and freeway authority officials in the identification of high-risk conditions, these methods have the potential to assist in the implementation of appropriate countermeasures.

The findings of the study suggest that future research in real-time crash likelihood prediction on freeways should focus on using hybrid deep learning models to evaluate the results of this study. The future research could further enhance predictive accuracy and performance of DL models by exploring more intricate architectures and structures, such as Transformer, gated recurrent network (GRU), and long-short term memory (LSTM).

## ACKNOWLEDGMENTS

## AUTHOR CONTRIBUTIONS

The authors confirm contribution to the paper as follows: study conception and design: S. Koohfar; data collection: S. Koohfar; analysis and interpretation of results: S. Koohfar; draft manuscript preparation: S. Koohfar. A. Kumar. All authors reviewed the results and approved the final version of the manuscript.